\documentclass{article}

    \usepackage[final,nonatbib]{neurips_2020}

\usepackage[utf8]{inputenc} %
\usepackage[T1]{fontenc}    %
\usepackage{hyperref}       %
\usepackage{url}            %
\usepackage{booktabs}       %
\usepackage{amsfonts}       %
\usepackage{nicefrac}       %
\usepackage{microtype}      %
\usepackage{color,soul}
\usepackage{amsmath}
\usepackage{graphicx}

\usepackage{xcolor,soul}

\title{Exploring Contrastive Learning in \\Human Activity Recognition for Healthcare}

\author{
  Chi Ian Tang$^1$, Ignacio Perez-Pozuelo$^{2,3}$, Dimitris Spathis$^1$, \textmd{and} Cecilia Mascolo$^1$ \\
  \\
  $^1$Department of Computer Science and Technology, University of Cambridge, UK\\
  $^2$MRC Epidemiology Unit, School of Clinical Medicine, University of Cambridge, UK\\
  $^3$Alan Turing Institute, UK
}

\begin{document}

\maketitle

\begin{abstract}
    Human Activity Recognition (HAR) constitutes one of the most important tasks for wearable and mobile sensing given its implications in human well-being and health monitoring.
    Motivated by the limitations of labeled datasets in HAR, particularly when employed in healthcare-related applications, this work explores the adoption and adaptation of \emph{SimCLR}, a contrastive learning technique for visual representations, to HAR. The use of contrastive learning objectives causes the representations of corresponding views to be more similar, and those of non-corresponding views to be more different. After an extensive evaluation exploring 81 combinations of different signal transformations for augmenting the data, we observed significant performance differences owing to the order and the function thereof. In particular, preliminary results indicated an improvement over supervised and unsupervised learning methods when using fine-tuning and random rotation for augmentation, however, future work should explore under which conditions \emph{SimCLR} is beneficial for HAR systems and other healthcare-related applications. 
\end{abstract} %
\section{Introduction}

The increasing adoption of mobile devices has resulted in vast amounts of data streams that can be used to better understand human behaviors and could have important implications for health and well-being.
Human activity recognition (HAR) is one of the fundamental tasks in mobile sensing with important implications in many healthcare applications \cite{health_har_avci2010activity}. Advancements in deep learning research have provided ground work for more complex and accurate HAR systems~\cite{deep_har_1d, deep_har_2, multi_self_har, deep_har_1, deep_har_survey_2}.

However, one of the major hurdles in applying data-oriented methods for digital health has been the difficulty in collecting labeled data. Sensor data streams from mobile devices are abstract and difficult to interpret by humans. As a result, labeled data collection usually require protocol-based tasks in laboratory settings~\cite{hhar, motionsense, unimib, uci_har, wisdm}. The fixed protocols adopted further limit the generalizability of these systems in healthcare applications. 

Unsupervised and semi-supervised learning techniques, which use unlabeled data as an alternative source of training data, have been proposed to overcome the limitations associated with the lack of labels~\cite{semi_supervised_survey, semi_supervised_intro}.
Contrastive learning, in which models learn to extract representations by contrasting positive pairs (samples which are deemed to be similar) against negative pairs, is an area that is actively studied in computer vision~\cite{simclr_chen2020simple, moco_he2020momentum, cpc_oord2018representation, cpc2_henaff2019data}. These methods have shown superior performance in visual representation learning, and are also known to be less limited in terms of generalizability of learned representations compared to manually designed tasks~\cite{simclr_chen2020simple}.

In this study\footnote{Code available at \url{https://github.com/iantangc/ContrastiveLearningHAR}}, we explore for the first time, to the best of our knowledge, 
the effectiveness of contrastive learning techniques %
on sensor-based HAR data. In particular, we adapted \emph{SimCLR}~\cite{simclr_chen2020simple}, to enable contrastive learning using wearable sensor data. We use a set of data augmentation techniques designed for sensor time-series %
in place of image augmentation operators and the technique is evaluated on open HAR datasets. A systematic analysis of the effect of different data transformation methods showed that different combinations of transformations can lead to significant differences in performance. The representations learned by the \emph{SimCLR} framework led to better performance compared to fully-supervised training when models are fine-tuned with labels. This work is indicative of the potential of contrastive learning frameworks in HAR, and due to the modality-agnostic nature of contrastive learning, this method can potentially be generalized to other types of data, opening the door to further ideas that fully leverage these techniques.  
\section{Related work}

Semi-supervised learning techniques have been studied by many researchers to utilize unlabeled data in increasing the diversity and quantity of data used for training \cite{semi_supervised_survey, semi_supervised_intro, semi_supervised_survey2, semi_har}. Within semi-supervised training techniques, self-supervised learning has been actively studied due to their similarity to supervised methods and their promising results. 

\subsection{Contrastive learning for visual representations}

Contrastive learning is proposed to capture the intrinsic structures of data without \textit{ad-hoc} heuristics. Recently, various methods were proposed by researchers to leverage unlabeled data, including \emph{Contrastive Predictive Coding} \cite{cpc_oord2018representation} and \emph{Momentum Constrast} (MoCo) \cite{moco_he2020momentum}.

Recent work by Chen et al. proposed a relatively straight-forward contrastive learning framework, \emph{SimCLR} \cite{simclr_chen2020simple}. In their work they use transformation functions to generate positive pairs, and then add a projection network before minimizing the contrastive loss. Compared to other methods, the use of transformed views of data samples within the same batch as negative pairs reduces the need for high usage of memory and allows flexible batch sizes. State-of-the-art performance was reported, and the performance gain when there is low availability of data is of particular interest.

\subsection{Human activity recognition}
Studies have pointed out the importance of HAR in healthcare due to its implications for pervasive and ubiquitous sensing \cite{health_har_avci2010activity, health_har_2_ogbuabor2018human, health_har_3_subasi2018iot}, however, thus far the scale of the labeled datasets have been limited by the difficulty of collection data of this nature. 

The application of self-supervised learning in wearable-based human activity recognition is less common than in computer vision. Saeed et al. proposed the use of transformation discrimination in which models are trained to identity which transformation has been applied to a particular sample, as an learning objective for HAR models \cite{multi_self_har}. A performance gain was reported compared to the fully-supervised approach. However, the use of a different, unlabeled dataset for training saw little to no improvement with this task.
\section{Method}

In this section, we first describe the \emph{SimCLR} contrastive learning framework and then illustrate the modifications which we adopted to explore its application to HAR.

\subsection{SimCLR}

The \emph{SimCLR} contrastive learning framework~\cite{simclr_chen2020simple} consists of four main components which are adopted in our design. The framework does not have underlying assumptions on the modality of data. 

\textbf{A probabilistic transformation function} which randomly transforms a data sample into a different view of the sample. During the generation of training data, this function is applied to the same data sample twice, resulting in two different views of the same data, which is then used as the positive pair for training.

\textbf{A neural network base encoder}, which is responsible for encoding the data samples into a latent space. 

\textbf{A projection head}, which is effectively another neural network that projects the representations in the latent space into another space for contrastive learning.

\textbf{A contrastive loss function} which defines the learning objective. In our case, the \emph{NT-Xent} (normalized temperature-scaled cross entropy loss) \cite{simclr_chen2020simple, nt_xent_sohn2016improved, nt_xent_wu2018unsupervised} was adopted.
    
\subsection{SimCLR for HAR}

As the \emph{SimCLR} framework was originally designed for visual representation learning \cite{simclr_chen2020simple}, modifications are necessary for adapting the framework for HAR.

\subsubsection{Transformation functions} 

In this study, eight time-series transformation functions, as used in \cite{multi_self_har} and \cite{har_transformations}, were utilized to compose the transformation function for contrastive learning:

1. \textbf{Adding random Gaussian noise.} Random noise signals with a mean of zero and a standard deviation of $-0.05$ are added to the data sample.

2. \textbf{Scaling by a random factor.} Each channel of the signal is scaled by a random factor that is drawn from a normal distribution of mean being $1.0$ and standard deviation being $0.1$.

3. \textbf{Applying a random 3D rotation.} A random axis in 3D and a random rotational angle are drawn with a uniform distribution, and the corresponding rotation is applied to the sample. 

4. \textbf{Inverting the signals.} The values of the sample are multiplied by a factor of $-1$.

5. \textbf{Reversing the direction of time.} The entire window of the sample is flipped in the time-direction.

6. \textbf{Randomly scrambling sections of the signal.} The signal is segmented into 4 different sections, and a random permutation is performed on the segments, and then recombined.

7. \textbf{Stretching and warping the time-series.} A random cubic spline with 4 fixed points is generated and this defines the deviation of the speed of time flow from normal. The signal is then stretched and warped according to the cubic spline.

8. \textbf{Shuffling the different channels.} The 3 channels of the signals are randomly permuted.

A subset of these functions is selected and applied in different orders to form the overall probabilistic transformation function.

\subsubsection{Other components} In this work, a relatively lightweight neural network architecture, \emph{TPN} (as proposed in~\cite{multi_self_har}) was adopted as the base encoder to suit the need for HAR systems. A three-layer fully connected MLP was used as the projection head, and the NT-Xent loss function was kept unchanged.

\section{Evaluation and results}

\subsection{Evaluation protocol}

In order to evaluate the effectiveness of the proposed framework, we conducted the evaluation with the following protocol.

\textbf{Dataset.} A publicly available dataset, MotionSense \cite{motionsense} was used in our evaluation. The dataset contains data collected from 24 subjects who carried an iPhone 6s in their trousers' front pocket while performing 6 different activities: walking downstairs, walking upstairs, walking, jogging, sitting, and standing. Data from the tri-axial accelerometer at 50 Hz was used in this study, forming $6630$ windows, each with 400 timestamps and 50\% overlap. 

\textbf{Evaluation metrics.} Two evaluation protocols were used in this study: linear evaluation and fine-tuned evaluation. The linear evaluation protocol follows that presented in \emph{SimCLR}, where only an additional fully connected layer is attached to the base encoder and trained~\cite{simclr_chen2020simple}. The fine-tuned evaluation protocol is similar to that in \cite{multi_self_har},
in which all layers of the base encoder are frozen except for the last one, and a two-layer fully connected classification head is attached and fine-tuned. The models are trained on data from 19 subjects and then evaluated on unseen data from the remaining 5 subjects. The F1 weighted scores on the activity labels are reported as the evaluation metric.

\textbf{Training setup.} The base encoder consists of three temporal (1D) convolutional layers, with kernel sizes of 24, 16, 8, and 32, 64 and 96 filters respectively. The ReLU activation function is used with a dropout rate of $0.1$. A global maximum pooling layer is added at the end. During pre-training, the projection head consists of 3 fully-connected layers with 256, 128 and 50 units respectively, and the classification head in fine-tuned evaluation consists of two fully connected layers, with 1024 and 6 units respectively. The SGD optimizer with a cosine decay of learning rate is used during pre-training for 200 epochs and a batch size of 512. For linear evaluation, the model is trained for 50 epochs with the SGD optimizer and a learning rate of 0.03. For fine-tuned evaluation, the model is fine-tuned with Adam optimizer and a learning rate of 0.001 for 50 epochs.

\subsection{Evaluation Results}

\begin{figure}
    \centering
    \includegraphics[width=0.92\textwidth]{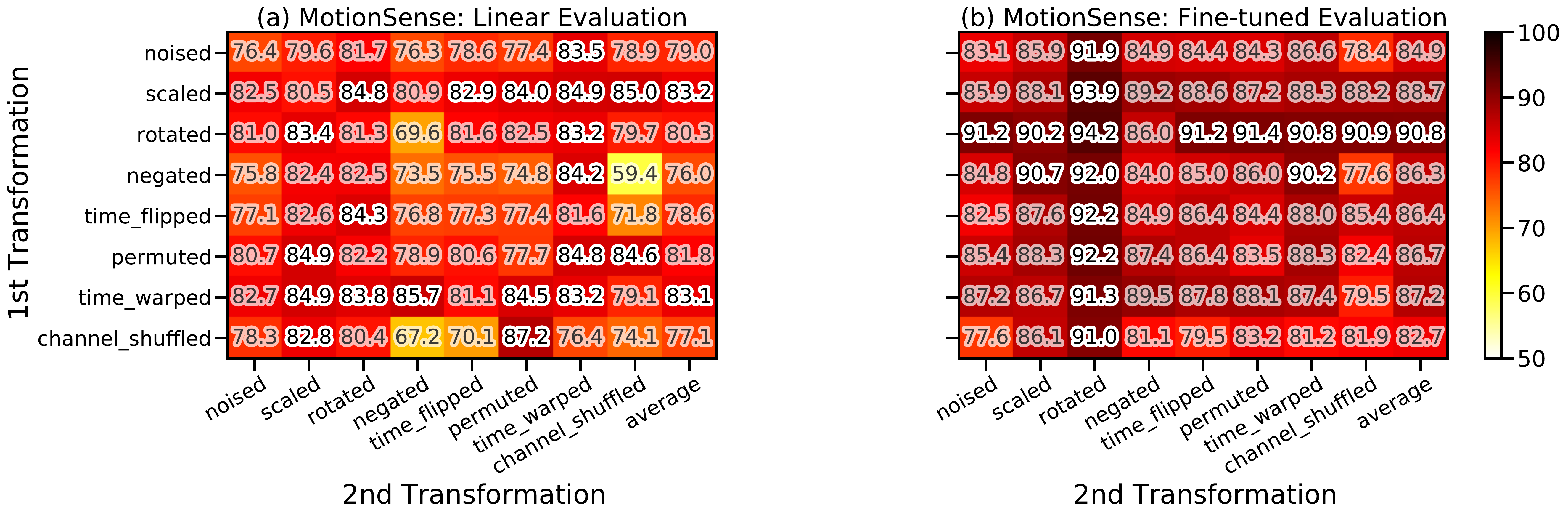}
    \caption{Average weighted F1 scores (in percent) of models trained by different combinations of transformation functions for \emph{SimCLR} on the MotionSense dataset across 5 independent runs. The diagonal entries correspond to using only a single transformation, and the last column is the average performance of the corresponding rows.}
    \label{fig:motion_sense_transform}
\end{figure}

In order to evaluate the impact of using different transformation for \emph{SimCLR} pre-training, linear and fine-tuned evaluations were performed on the MotionSense dataset. Figure~\ref{fig:motion_sense_transform} shows the average performance of the models pre-trained by \emph{SimCLR}. In linear evaluation, we observed that when using the scaled and the time-warped transformations generally performed well when used alongside other transformations. However, the highest performing models were trained by combining channel shuffling and permutation, with an average F1 score of $0.872$. When the models where further fine-tuned (see figure \ref{fig:motion_sense_transform} (b)), the rotation transformation out-performed all other transformation functions, achieving the highest F1 score of $0.942$ when used alone. Some combinations perform significantly worse, where the performance difference can be as high as $0.278$ for the linear evaluation.

We also trained models using pure supervised learning and reproduced results from previous work in self-supervised learning for HAR (see \cite{multi_self_har}), in which the models are pre-trained to identify transformation rather than optimizing the contrastive objective. The average F1 scores are $0.922$ and $0.923$ respectively. Our adaptation of \emph{SimCLR} for HAR resulted in a performance gain of up to $0.020$ compared to fully supervised models.  
\section{Discussion and Conclusion}

In this work, we adapted the \emph{SimCLR} contrastive learning framework from visual representation learning to HAR, one of the most relevant tasks for digital health applications. In our evaluation, we observed that the \emph{SimCLR} framework displays promising results, slightly outperforming other fully-supervised and semi-supervised methods, which is indicative of the potential of transferring \emph{SimCLR} to mobile sensing settings and other health data, especially due to the modality-agnostic nature of the method.  We also observed that the use of different transformation functions can affect the performance of the models, and in some cases, to a significant degree. The relationship between performance and transformation functions is not consistent across different evaluation protocols, and further work involving more evaluation datasets and transformations will help shed light on this issue.

\begin{ack}
This work is partially supported by Nokia Bell Labs through their donation for the Centre of Mobile, Wearable Systems and Augmented Intelligence to the University of Cambridge. CI.T is additionally supported by the Doris Zimmern HKU-Cambridge Hughes Hall Scholarship and from the Higher Education Fund of the Government of Macao SAR, China. D.S is supported by the Embiricos Trust Scholarship of Jesus College Cambridge, and EPSRC through Grant DTP (EP/N509620/1). I.P is supported by GlaxoSmithKline and EPSRC through an iCase fellowship (17100053). The authors declare that they have no conflict of interest with respect to the publication of this work.
\end{ack} 
\bibliographystyle{unsrt}
\bibliography{arxiv_main}

\end{document}